\documentclass[letterpaper]{article}
\PassOptionsToPackage{table}{xcolor}
\usepackage{aaai2027}
\usepackage[hyphens]{url}
\usepackage{graphicx}
\usepackage{natbib}
\usepackage{caption}
\usepackage{amsmath}
\usepackage{amssymb}
\usepackage{booktabs}
\usepackage{multirow}
\usepackage{array}
\usepackage{makecell}
\newcommand{\method}{WaveTLM}
\newcommand{\chatts}{ChatTS}

\newcommand{\exectsqa}{ExecTS-QA}
\newcommand{\scitstimeomni}{SciTS}

\newcommand{\best}[1]{\textbf{#1}}
\newcommand{\dash}{\textcolor{gray}{--}}
\newcommand{\yes}{\checkmark}
\newcommand{\no}{$\times$}
\newcommand{\partialmark}{$\triangle$}
\newcolumntype{C}[1]{>{\centering\arraybackslash}m{#1}}
\newcolumntype{L}[1]{>{\raggedright\arraybackslash}m{#1}}
\definecolor{oursgreen}{HTML}{EAF6EA}
\definecolor{softgray}{HTML}{F5F6F8}

\title{WaveTLM: Reliable Time-Series Language Modeling through Task Compilation}
\author{
    Jiahui Chen\textsuperscript{\rm 1},
    Bingke Zhu\textsuperscript{\rm 1},
    Hongyu Pan\textsuperscript{\rm 1},
    Yingying Chen\textsuperscript{\rm 1}
}

\affiliations{
    \textsuperscript{\rm 1}Institute of Automation,
    Chinese Academy of Sciences\\
    Beijing, China
}

\begin{document}

\maketitle

\begin{abstract}
Time-series language models provide a shared natural-language interface
across temporal tasks, but plausible text does not guarantee reliable task
outputs. Responses may appear reasonable while hallucinating the required
object: numerical sequences can violate shape, scale, channel order, or
temporal alignment, and textual decisions can fall outside the legal label
space. We formulate \emph{reliable time-series language modeling},
separating task-object reliability from predictive quality. We introduce
\exectsqa{}, a contract-grounded benchmark spanning forecasting, imputation,
classification, anomaly detection, and waveform analysis. We further propose
\method{}, a unified compiler-executor model whose task compiler transforms
user requests, visible arguments, and wave-grounded evidence into typed task
states, while task-native executors construct numerical tensors, legal
decisions, or structured records. On \exectsqa{}, a single \method{}
checkpoint achieves 99.40\% contract-valid coverage, compared with 37.83\%
for the strongest evaluated string-first baseline, while retaining balanced
predictive performance across all five task families. Evaluations on SciTS,
TSQA, IRTS-ToolBench, and ARFBench provide additional evidence of transfer.
The code, construction scripts, and \exectsqa{} dataset will be publicly
released upon publication. These results show that task compilation can
convert plausible language generation into reliable time-series outputs.
\end{abstract}

\section{Introduction}

\begin{figure}[!t]
\centering
\IfFileExists{task_native_execution.pdf}{
\includegraphics[width=0.86\columnwidth]{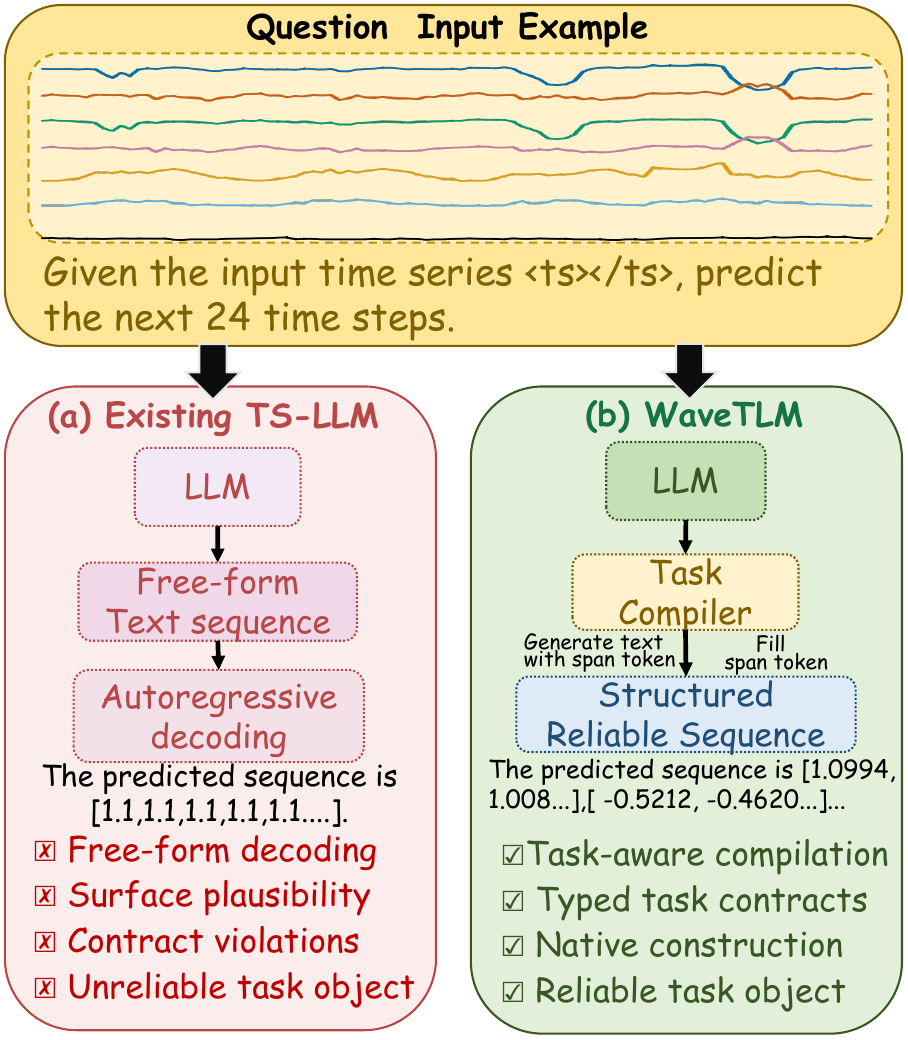}
}{
\fbox{\parbox[c][0.30\textheight][c]{0.84\columnwidth}{\centering
\textbf{Figure placeholder: task\_native\_execution.pdf.}}}
}
\caption{\textbf{From plausible text to reliable time-series outputs.}
Existing TS-LLMs generate free-form text, whereas \method{} compiles task
intent and temporal evidence into a typed state and constructs a
contract-aligned output.}
\label{fig:task_native_execution}
\end{figure}

Natural-language interfaces provide shared access to heterogeneous
time-series tasks
~\cite{promptcast,llmtime,autotimes,gpt4ts,test,timellm,unitime,chatts,
timemqa,timemaster,timeomni,insightminer,scits}. However, plausible language
does not guarantee a reliable task output. Free-form forecasts may violate
the required horizon, channel order, scale, or temporal alignment, while
classification and anomaly responses may be ambiguous or outside the legal
label space. We call this failure \emph{task-object hallucination}: the text
appears reasonable but fails to instantiate the numerical tensor, legal
decision, or structured record required by the task. As illustrated in
Figure~\ref{fig:task_native_execution}, existing string-first systems rely on
free-form generation followed by output parsing, whereas \method{} explicitly
binds the request to a typed contract before constructing the required object.
The figure summarizes the central motivation of our work and clarifies why
linguistic plausibility and task-object reliability must be evaluated
separately. It also highlights that output reliability depends on the model's
construction path, rather than only on the surface form of its response.

Task-object reliability requires the complete output to satisfy its type,
shape, alignment, scale, channel order, finite-value constraints, and
decision domain. It is distinct from predictive quality: an inaccurate but
valid forecast remains measurable, whereas an invalid sequence does not.
Surface controls such as JSON schemas, grammar-constrained decoding, or
iterative correction may improve parseability, but do not determine horizon,
mask alignment, inverse normalization, channel semantics, or legal labels.
The key difficulty is that these constraints are jointly determined by the
request, temporal input, and task protocol. The same observed sequence may
require a future tensor for forecasting, mask-aligned values for imputation,
or a single legal decision for classification. A language-facing model must
therefore recover not only plausible content, but also the requested task,
its admissible output space, and the structural correspondence between input
and response. Leaving these requirements implicit in free-form decoding makes
them difficult to enforce and verify.

We therefore formulate reliable time-series language modeling as compilation
followed by execution.\footnote{We use \emph{execution} in the
compiler-system sense: a task compiler maps the request, visible arguments,
and temporal evidence to a typed state, while an executor instantiates the
corresponding task-native object under its registered contract.}
The compiler identifies the requested operation, binds visible arguments to
an output contract, and extracts task-relevant evidence. The executor then
constructs the required object instead of autoregressively generating every
value or decision. This separation makes the output type and construction
path explicit before prediction while retaining language generation for task
access and explanation. Reliability thus follows from the model path and task
contract rather than post-hoc repair.

We introduce \exectsqa{}, a contract-grounded evaluation suite spanning five
task families. Each instance pairs a natural-language request and task-native
target with a deterministic contract specifying the admissible type,
dimensions, alignment, scale, channel order, and decision domain. Validation
applies no evaluator-side truncation, reshaping, scale correction, or label
remapping. Supplementary Figure~S1 summarizes the benchmark composition,
showing the five task families, their heterogeneous source domains, and the
corresponding evaluator-facing objects. It complements
Figure~\ref{fig:task_native_execution} by connecting the conceptual reliability
problem to the concrete tasks evaluated in \exectsqa{}. Together, the two
figures connect the motivating failure mode with the benchmark structure used
to measure it across heterogeneous tasks. We further propose \method{}, whose
task compiler combines the request, semantic temporal patches, and multi-scale
waveform evidence into typed task states. Task-native executors then construct
numerical tensors, legal decisions, and structured waveform records.

\begin{table}[!t]
\centering
\scriptsize
\renewcommand{\arraystretch}{1.06}
\setlength{\tabcolsep}{1.7pt}
\begin{tabular}{
L{0.205\columnwidth}
C{0.125\columnwidth}
C{0.125\columnwidth}
C{0.145\columnwidth}
C{0.145\columnwidth}
C{0.145\columnwidth}}
\toprule
\textbf{Method}
& \makecell{\textbf{Lang.}\\\textbf{query}}
& \makecell{\textbf{Multi-}\\\textbf{task}}
& \makecell{\textbf{Typed}\\\textbf{contracts}}
& \makecell{\textbf{Native}\\\textbf{construction}}
& \makecell{\textbf{Strict}\\\textbf{validation}} \\
\midrule

UniTS
& \no
& \yes
& \no
& \yes
& \no \\

Time-LLM
& \partialmark
& \no
& \no
& \no
& \no \\

\chatts{}
& \yes
& \partialmark
& \no
& \no
& \no \\

Time-MQA
& \yes
& \yes
& \no
& \no
& \no \\

TimeOmni-1
& \yes
& \partialmark
& \no
& \no
& \no \\

\scitstimeomni{}
& \yes
& \yes
& \no
& \partialmark
& \no \\

\textbf{\method{}}
& \yes
& \yes
& \yes
& \yes
& \yes \\

\bottomrule
\end{tabular}
\caption{\textbf{Language-facing and reliability-oriented capabilities.}
Native construction instantiates task objects through dedicated numerical or
decision paths. $\triangle$ denotes partial support.}
\label{tab:capabilities}
\end{table}

As shown in Table~\ref{tab:capabilities}, \method{} combines a shared language
interface with typed contracts, native construction, and strict validation.
A single checkpoint achieves 99.40\% contract-valid coverage on
\exectsqa{}, compared with 37.83\% for the strongest evaluated string-first
baseline, while retaining competitive task-dependent utility. Evaluations on
SciTS, TSQA, IRTS-ToolBench, and ARFBench provide additional transfer
evidence.

\begin{samepage}
The main contributions are:
\begin{itemize}
    \item \textbf{Reliable TLM formulation.}
    We identify task-object hallucination as a reliability problem and
    distinguish task-object reliability from predictive quality.

    \item \textbf{Contract-grounded evaluation.}
    We introduce \exectsqa{}, a five-task evaluation suite with task-native
    targets, deterministic contracts, and strict validation without
    evaluator-side repair.

    \item \textbf{Compiler-executor model.}
    We propose \method{}, which compiles requests and wave-grounded evidence
    into typed task states and constructs reliable numerical, decision, and
    structured outputs through task-native executors.
\end{itemize}
\end{samepage}
\section{Related Work}

\paragraph{Native temporal modeling.}
Task-specific architectures directly construct numerical tensors under fixed
temporal interfaces~\cite{deepar,nbeats,tcn,informer,autoformer,fedformer,
nonstationary,patchtst,itransformer,timesnet}. Foundation and multitask models
extend this paradigm through pretrained temporal representations, probabilistic
forecasting, and shared task architectures
~\cite{units,chronos,timesfm,moment,moirai,lagllama,timegpt,timefound,
timedit,tsicl}. Their output heads provide useful structural guarantees, but
they generally assume task-specific inputs rather than inferring the required
operation and object from a natural-language request. \method{} retains native
object construction while placing it behind a request-derived task compiler.

\paragraph{Language models for time series.}
Prior work adapts pretrained language models as temporal backbones,
representation learners, or forecasters
~\cite{gpt4ts,timellm,autotimes,unitime}, and language-facing systems expose
prediction and understanding through prompting or shared QA interfaces
~\cite{promptcast,llmtime,chatts,timemqa,timemaster,timeomni,
insightminer,scits}. These studies establish the value of semantic priors and
heterogeneous task access. Their published protocols, however, primarily
measure prediction error, answer accuracy, or response-level quality; whether
the complete response instantiates the evaluator's required native object is
not usually isolated as a full-denominator reliability property. \method{}
separates this structural reliability from task-specific predictive utility.

\paragraph{Time-series QA and reliability evaluation.}
TSQA, SciTS, IRTS-ToolBench, and ARFBench broaden temporal evaluation through
common question-answering or reasoning protocols
~\cite{timemqa,scits,irts_toolbench,arfbench}. \exectsqa{} is complementary:
it tests whether a language-facing model produces a contract-valid future
tensor, mask-aligned value set, legal decision, or structured waveform record
without evaluator-side repair. The supplement provides an expanded comparison
of these lines of work and clarifies the distinction in output construction
and evaluation protocol.

\section{Method}

\begin{figure*}[t]
\centering
\IfFileExists{WaveTLM-Overview.pdf}{
\includegraphics[width=1.0\textwidth]{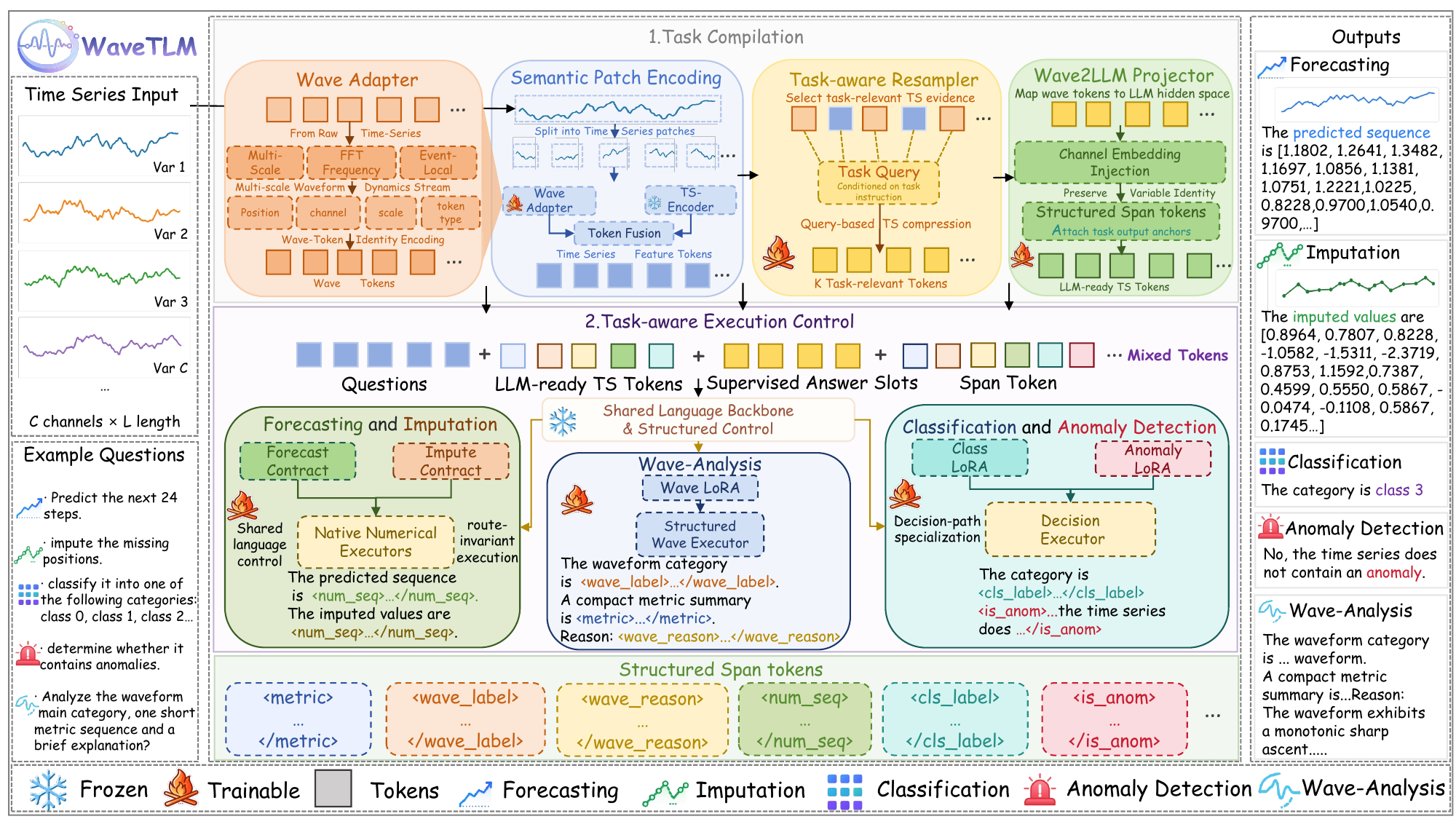}
}{
\fbox{\parbox[c][0.25\textheight][c]{0.95\textwidth}{\centering
\textbf{Figure placeholder: WaveTLM-Overview.pdf.} The final figure
should distinguish the task compiler, task-native executors, semantic
and raw-wave evidence, and typed output states. Frozen and trainable
modules should use different visual styles.}}
}
\caption{\textbf{Compiler-executor architecture of \method{}.}
The task compiler binds a visible request to a typed contract and compiles
semantic and waveform evidence into a task-conditioned state. Task-native
executors then construct the corresponding numerical tensor, legal decision,
or structured temporal record.}
\label{fig:overview}
\end{figure*}

Figure~\ref{fig:overview} provides the architectural roadmap for the method.
From left to right, \method{} encodes semantic and raw-wave evidence, compiles
a route-conditioned typed state under the selected contract, and activates
the corresponding task-native executor. The following subsections formalize
this request-to-contract-to-object pipeline.

\subsection{Reliable TLM Formulation}

Given a multivariate series $X\in\mathbb{R}^{C\times L}$, a
natural-language request $q$, and visible task arguments $m$, \method{}
infers the requested task family
\begin{equation}
\widehat{\tau}=R(q),
\qquad
\widehat{\tau}\in\mathcal{T},
\end{equation}
where $\mathcal{T}$ comprises forecasting, imputation, classification,
anomaly detection, and waveform analysis. Routing uses only the visible
request; $m$ may specify the horizon, observation mask, channel count,
label vocabulary, or normalization state, but never the target.

Each task family $\tau$ has a deterministic contract
\begin{equation}
\mathcal{C}_{\tau}(m)=
\bigl(
\mathcal{Y}_{\tau}(m),
\pi_{\tau},
\nu_{\tau},
\sigma_{\tau}
\bigr),
\end{equation}
where $\mathcal{Y}_{\tau}(m)$ is the admissible task-object space,
$\pi_{\tau}$, $\nu_{\tau}$, and $\sigma_{\tau}$ are the registered parser,
validator, and serializer. The contract specifies object type, shape,
alignment, scale, channel order, finite-value constraints, and legal
decision domain.

The compiler-executor path is
\begin{equation}
\begin{aligned}
h_{\widehat{\tau}}
&=\operatorname{Comp}_{\widehat{\tau}}(q,X,m),\\
z_{\widehat{\tau}}
&=\operatorname{Exec}_{\widehat{\tau}}
(h_{\widehat{\tau}},X,m)
\in\mathcal{Y}_{\widehat{\tau}}(m),\\
y&=\sigma_{\widehat{\tau}}(z_{\widehat{\tau}}),
\qquad
\overline z_{\widehat{\tau}}=\pi_{\widehat{\tau}}(y).
\end{aligned}
\end{equation}
Here, $h_{\widehat{\tau}}$ is the typed intermediate state,
$z_{\widehat{\tau}}$ is the native task object, and
$\overline z_{\widehat{\tau}}$ is its round-trip recovery from the rendered
response. Parsing failures are treated as invalid objects.

Task-object reliability is defined as
\begin{equation}
\begin{aligned}
r_{\tau}
={}&
\mathbb{1}[\widehat{\tau}=\tau]\,
\mathbb{1}[\nu_{\tau}(z_{\widehat{\tau}};m)=1]\\
&\cdot
\mathbb{1}\!\left[
\nu_{\tau}(\overline z_{\widehat{\tau}};m)=1
\land
\overline z_{\widehat{\tau}}
\approx_{\tau}
z_{\widehat{\tau}}
\right],
\end{aligned}
\end{equation}
where $\approx_{\tau}$ denotes the registered round-trip equivalence
relation. Reliability is measured over the full request denominator, while
task-specific metrics separately evaluate the predictive quality of
contract-valid objects. No incorrect route or invalid object is repaired
using evaluator-side information.

\subsection{Task Compiler}

The task compiler determines the required object and extracts the temporal
evidence needed to construct it. It combines semantic patches, raw waveform
features, task-aware resampling, and contract binding.

The semantic encoder and Wave Adapter produce complementary representations:
\begin{equation}
\begin{aligned}
S&=f_{\mathrm{sem}}(X),\\
Z_{\mathrm{raw}}&=G_{\mathrm{wave}}(X),\\
Z&=
F_{\mathrm{wave}}
\left(
[S';Z_{\mathrm{raw}}]+E_{\mathrm{meta}}
\right).
\end{aligned}
\end{equation}
Here, $S$ contains language-aligned temporal patches, while
$Z_{\mathrm{raw}}$ preserves multi-scale temporal, frequency-domain,
event-local, and channel-aware evidence. $E_{\mathrm{meta}}$ retains
structural identities such as channel, position, scale, and token type.

Because different tasks require different evidence from the same sequence,
a route-conditioned resampler produces a task-specific representation:
\begin{equation}
\begin{aligned}
Q_{\widehat{\tau}}
&=
Q_{\widehat{\tau}}^{(0)}
+f_q(q)+f_m(m),\\
\widetilde Z_{\widehat{\tau}}
&=
\operatorname{CrossAttn}
(Q_{\widehat{\tau}},Z,Z).
\end{aligned}
\end{equation}
The resulting compiler state is
\begin{equation}
h_{\widehat{\tau}}
=
\left(
\widetilde Z_{\widehat{\tau}},
a_{\widehat{\tau}},
\mathcal{C}_{\widehat{\tau}}(m)
\right),
\end{equation}
where $a_{\widehat{\tau}}$ is the selected typed answer field. Thus, the
compiler jointly specifies the task-relevant evidence, admissible output
space, and executor path without directly predicting the final object.
Detailed Wave Adapter branches, metadata embeddings, and resampler
configurations are provided in the supplement.
\subsection{Task-Native Executors}

The compiled contract selects an executor for one of three output forms:
continuous numerical objects, discrete decisions, or structured waveform
records.

\paragraph{Continuous objects.}
Forecasting and imputation do not autoregressively generate every scalar.
Instead, a numerical executor constructs
\begin{equation}
\widehat{Y}_{\widehat{\tau}}
=
D_{\widehat{\tau}}
\left(
X,M,Z_{\mathrm{num}},m
\right),
\end{equation}
where $M$ is the observation mask when applicable and
$Z_{\mathrm{num}}$ is the compiled numerical representation. For
forecasting, the contract determines the prediction horizon and channel
organization. For imputation, it determines the exact correspondence
between predictions and missing positions. Reversible normalization is
inverted before serialization, so the values remain learned predictions
while their shape, alignment, channel order, and scale state follow the
compiled task path.

Typed answer fields and numerical executors serve complementary roles.
The typed field marks where the required object appears in the language
response, while the executor constructs its numerical content from the
observed sequence and compiled evidence.

\paragraph{Decision and structured objects.}
Classification and anomaly detection use specialized decision paths whose
outputs are restricted to their registered label domains. Waveform analysis
uses separate categorical, numerical, and explanatory fields: task-specific
heads construct the structured attributes, while the language model produces
the associated explanation. This separation allows object reliability and
predictive quality to be evaluated independently.

Every executor output undergoes native contract validation, serialization,
and round-trip recovery. A response is contract-valid only when both the
native and recovered objects satisfy the registered contract and remain
equivalent. Validation never truncates, pads, reshapes, rescales, remaps, or
otherwise modifies the prediction. Detailed executor definitions, typed-field
schemas, and validation algorithms are included in the supplement.

\subsection{Multitask Learning and Inference}

The language backbone and pretrained semantic patch encoder remain frozen.
Trainable components include the Wave Adapter, route-conditioned resampler,
typed-output embeddings, task-native executors, and task-specific LoRA paths
for language-mediated tasks. Continuous tasks rely primarily on their
numerical executors, whereas classification, anomaly detection, and waveform
analysis additionally use specialized language paths.

The multitask objective is
\begin{equation}
\mathcal{L}
=
\mathcal{L}_{\mathrm{LM}}
+
\lambda_{\mathrm{slot}}\mathcal{L}_{\mathrm{slot}}
+
\sum_{\tau\in\mathcal{T}}
g_{\tau}\lambda_{\tau}\mathcal{L}_{\tau},
\end{equation}
where $g_{\tau}$ activates the loss of the current task family.
$\mathcal{L}_{\mathrm{LM}}$ supervises language generation,
$\mathcal{L}_{\mathrm{slot}}$ supervises typed answer fields, and
$\mathcal{L}_{\tau}$ denotes the corresponding numerical, decision, or
structured-output objective. Task-specific loss definitions and optimization
settings are reported in the supplement.

At inference, the model receives only the visible request, time series, and
task arguments. The request-derived route activates the corresponding
contract, compiled representation, typed answer field, and executor. No
evaluator-side task label, route correction, output repair, scale correction,
or label remapping is used; a reliable object must therefore be produced by
the selected compiler-executor path itself.
\section{Experiments}

\subsection{Experimental Setup}

\exectsqa{} is a contract-grounded benchmark containing 35,323 training
instances and 4,795 evaluation instances: 515 forecasting, 261 imputation,
868 classification, 2,592 anomaly-detection, and 559 waveform-analysis
examples. Each instance combines a natural-language request, visible task
arguments, a task-native target, and a deterministic output contract
specifying the required type, shape, temporal alignment, scale, channel
order, and legal decision domain.

We compare \method{} with Qwen3-8B-SFT, Raw ChatTS, and ChatTS-SFT under the
same visible requests and contract validators. UniTS is included as a native
numerical reference for forecasting and imputation, but is excluded from
language-interface coverage comparisons because it receives task-specific
native inputs rather than natural-language requests. \method{} uses a frozen
Qwen3-8B language backbone and the pretrained temporal encoder from
\chatts{}. The Wave Adapter, route-conditioned resampler, task-native
executors, typed-output embeddings, and task-specific LoRA paths are
trainable. Full implementation and optimization details are provided in the
supplement.

External evaluation includes SciTS~\cite{scits},
TSQA~\cite{timemqa}, IRTS-ToolBench~\cite{irts_toolbench}, and
ARFBench~\cite{arfbench}. The central checkpoint is evaluated without
benchmark-specific training on SciTS, IRTS-ToolBench, and ARFBench, whereas
TSQA follows its official supervised adaptation protocol with 1,000 examples
per task. All benchmarks retain their published task definitions and metrics.

The experiments examine whether \method{} reliably instantiates the
requested task objects, whether contract-valid objects retain predictive
utility, whether the learned representations transfer beyond
\exectsqa{}, and whether compiler-executor components contribute through
distinct mechanisms.

We report Response Coverage (RC), Candidate Coverage (CC), Contract-Valid
Coverage (CVC), and Reliability Yield
$\mathrm{RY}=\mathrm{CVC}/\mathrm{CC}$. CVC requires every applicable
contract check to pass. Invalid, missing, or ambiguous outputs remain in the
full denominator, and no post-hoc truncation, padding, reshaping, scale
correction, or label remapping is applied. Task-specific predictive metrics
are reported separately on contract-valid objects.

\subsection{Task-Object Reliability}

Table~\ref{tab:execution_reliability} distinguishes returning a response,
recovering a candidate object, and satisfying the complete task-native
contract under identical requests and validators.

\begin{table}[!t]
\centering
\small
\renewcommand{\arraystretch}{1.12}
\setlength{\tabcolsep}{0pt}
\begin{tabular*}{\columnwidth}{@{\extracolsep{\fill}}lccccc@{}}
\toprule
& \multicolumn{3}{c}{\textbf{Coverage (\%)}}
& \multicolumn{2}{c}{\textbf{Reliability outcome}} \\
\cmidrule(lr){2-4}\cmidrule(l){5-6}
\textbf{Model}
& \textbf{RC} $\uparrow$
& \textbf{CC} $\uparrow$
& \textbf{CVC} $\uparrow$
& \textbf{RY} $\uparrow$
& \textbf{Gap} $\downarrow$ \\
\midrule

Qwen3-8B-SFT
& 8.47 & 7.09 & 0.02 & 0.29 & \underline{7.07} \\

Raw ChatTS$^{\dagger}$
& \underline{54.24} & 34.41 & 20.29 & 58.97 & 14.12 \\

ChatTS-SFT$^{\dagger}$
& \underline{54.24} & \underline{49.23} & \underline{37.83}
& \underline{76.84} & 11.40 \\

\textbf{\method{}}
& \best{99.98} & \best{99.40} & \best{99.40}
& \best{100.00} & \best{0.00} \\

\bottomrule
\end{tabular*}
\caption{\textbf{Stage-wise task-object reliability on \exectsqa{} (\%).}
All models receive the same visible requests and are evaluated under the
same strict contracts without post-hoc repair. CVC denotes Contract-Valid
Coverage, RY is $\mathrm{CVC}/\mathrm{CC}$, and Gap is
$\mathrm{CC}-\mathrm{CVC}$. Bold and underline denote the best and
second-best values; $^{\dagger}$ denotes a free-text interface.}
\label{tab:execution_reliability}
\end{table}

Raw ChatTS and ChatTS-SFT achieve only 20.29\% and 37.83\% CVC,
respectively. Fine-tuning improves both candidate recovery and reliability
yield, but 45.76\% of the evaluation instances still produce no response,
while 10.43\% produce illegal or ambiguous decisions. These are not
conventional prediction errors: the generated response fails to instantiate
the object on which the registered task metric is defined.

In contrast, \method{} aligns candidate and contract-valid coverage at
99.40\%, yielding 100\% reliability among recovered candidates. It constructs
contract-valid objects for all forecasting (515/515), imputation (261/261),
classification (868/868), and anomaly-detection (2,592/2,592) examples, and
for 530 of 559 waveform-analysis examples. Overall, the single checkpoint
produces 4,766 valid objects from 4,795 evaluation instances. The remaining
cases consist of one runtime failure and 28 waveform-routing failures.
Complete first-failure counts and task-wise distributions are reported in
the supplement.

A numerical-tolerance audit further examined 77 imputation objects initially
flagged by exact normalized-space comparison. After the registered inverse
transformation, the maximum discrepancy between native and round-trip objects
is $4.68\times10^{-7}$. These cases are therefore treated as
evaluator-equivalent floating-point representations rather than failures of
scale restoration or object reliability.

\subsection{Task-Specific Predictive Utility}

Task-object reliability establishes whether the requested object can be
evaluated, but not whether its prediction is accurate.
Table~\ref{tab:main_results} therefore reports one primary task-specific
metric for each task family, conditioned on contract-valid outputs.

\begin{table}[!t]
\centering
\small
\renewcommand{\arraystretch}{1.12}
\setlength{\tabcolsep}{0pt}
\begin{tabular*}{\columnwidth}{@{\extracolsep{\fill}}lccccc@{}}
\toprule
& \multicolumn{2}{c}{\textbf{Continuous}}
& \multicolumn{2}{c}{\textbf{Decision}}
& \textbf{Waveform} \\
\cmidrule(lr){2-3}\cmidrule(lr){4-5}\cmidrule(l){6-6}
\textbf{Model}
& \makecell{\textbf{Fcst.}\\MAE $\downarrow$}
& \makecell{\textbf{Imp.}\\MAE $\downarrow$}
& \makecell{\textbf{Cls.}\\Acc. $\uparrow$}
& \makecell{\textbf{Anom.}\\F1 $\uparrow$}
& \textbf{Cat.} $\uparrow$ \\
\midrule

Raw ChatTS$^{\dagger}$
& \dash{} & \dash{} & \underline{.167} & .368 & .558 \\

ChatTS-SFT$^{\dagger}$
& \dash{} & .769 & .086 & \best{.933} & \underline{.570} \\

UniTS$^{\ddagger}$
& \best{.311} & \best{.311} & N/A & N/A & N/A \\

\textbf{\method{}}
& \underline{.337} & \underline{.427} & \best{.706}
& \underline{.847} & \best{.713} \\

\bottomrule
\end{tabular*}
\caption{\textbf{Predictive utility of contract-valid objects on
\exectsqa{}.}
Metrics are computed only after the complete task contract is satisfied.
$^{\dagger}$ denotes a free-text interface, while UniTS$^{\ddagger}$
receives task-specific native inputs without language routing.
\dash{} indicates that no valid object is available. Complete metrics and
dataset-level results are reported in the supplement.}
\label{tab:main_results}
\end{table}

Among the evaluated language-facing systems, \method{} is the only model
with measurable task-specific utility across all five task families. UniTS
achieves lower forecasting and imputation MAE using task-specific native
inputs, while \method{} remains competitive through a shared
natural-language interface. For decision tasks, \method{} obtains the
highest classification accuracy and waveform-category accuracy.

ChatTS-SFT achieves higher anomaly F1 on the subset for which it returns a
valid decision. However, its 37.83\% overall CVC leaves a substantial
fraction of the evaluation set without a valid task object. This contrast
shows why predictive utility and task-object reliability must be reported
separately: a strong conditional score does not imply reliable task
completion over the full request distribution.

Tables~\ref{tab:execution_reliability} and~\ref{tab:main_results} therefore
measure complementary properties. The former evaluates whether the requested
object is produced over the full denominator, whereas the latter evaluates
prediction quality after the object has been reliably constructed. Because
the five task families use heterogeneous metrics, their predictive utility
is not collapsed into a single aggregate score.

\subsection{External Transfer}

The external benchmarks do not share the complete \exectsqa{} contracts and
are evaluated under their official protocols. They therefore provide
complementary evidence that the learned temporal representations and compiled
task paths remain useful beyond the benchmark on which task-object
reliability is measured. Results are reported without normalizing or pooling
scores across protocols.

\begin{table*}[!t]
\centering
\small
\renewcommand{\arraystretch}{1.18}
\setlength{\tabcolsep}{2.6pt}
\begin{tabular}{@{}
L{0.120\textwidth}
L{0.205\textwidth}
L{0.170\textwidth}
L{0.315\textwidth}
C{0.120\textwidth}
@{}}
\toprule
\textbf{Benchmark}
& \textbf{Model}
& \textbf{Numerical / Overall}
& \textbf{Decision / Other}
& \textbf{Setting} \\
\midrule

\multicolumn{5}{@{}l}{
\textbf{TSQA-50: supervised public-protocol adaptation}
} \\
\midrule

\multirow{6}{*}{TSQA-50}
& GPT-4o
& F: 1.790 / I: .018
& C: .320 / A: .640
& \multirow{6}{*}{task adapt.} \\

& Llama-3 8B
& F: 2.010 / I: .020
& C: .240 / A: .540
& \\

& Qwen-2.5 7B
& F: 1.820 / I: .016
& C: .520 / A: .680
& \\

& Mistral 7B
& F: 1.350 / I: \underline{.014}
& C: .440 / A: .580
& \\

& ChatTS-SFT
& F: \underline{1.018} / I: .178
& C: \underline{.760} / A: \underline{.740}
& \\

& \textbf{\method{}}
& F: \best{.072} / I: \best{.013}
& C: \best{.900} / A: \best{.820}
& \\

\midrule
\multicolumn{5}{@{}l}{
\textbf{IRTS-ToolBench: zero-shot irregular time-series QA}
} \\
\midrule

\multirow{3}{*}{IRTS-ToolBench}
& Qwen3.5-4B
& Overall: 55.18
& A.D.: 42.40 / CLS: \best{98.67} / TR: 46.00
& \multirow{3}{*}{zero-shot} \\

& DeepSeek-V4-Flash
& Overall: \underline{60.29}
& A.D.: \best{59.20} / CLS: 90.67 / TR: \underline{46.67}
& \\

& \textbf{\method{}}
& Overall: \best{63.29}
& A.D.: \underline{55.20} / CLS: \underline{94.67} /
TR: \best{53.33}
& \\

\midrule
\multicolumn{5}{@{}l}{
\textbf{ARFBench: zero-shot anomaly-oriented QA}
} \\
\midrule

\multirow{4}{*}{ARFBench}
& Random choice
& Overall: 24.50
& Tier I: 50.00 / weighted F1: \underline{22.50}
& -- \\

& OpenTSLM
& Overall: 0.80
& Tier I: 0.00 / weighted F1: 1.20
& \multirow{3}{*}{zero-shot} \\

& ChatTS
& Overall: \underline{31.10}
& Tier I: \underline{59.50} / weighted F1: 22.10
& \\

& \textbf{\method{}}
& Overall: \best{36.13}
& Tier I: \best{81.98} / weighted F1: \best{24.83}
& \\

\midrule
\multicolumn{5}{@{}l}{
\textbf{SciTS: protocol-compatible scientific transfer}
} \\
\midrule

& \textbf{Model}
& \multicolumn{1}{c}{
\makecell{\textbf{Numerical SR} $\uparrow$\\$(N=4{,}019)$}
}
& \multicolumn{1}{c}{
\makecell{\textbf{Anomaly F1} $\uparrow$\\
\textbf{MEU01 / PHU04 / PHU05 / URU04}}
}
& \textbf{Setting} \\
\cmidrule(lr){2-2}
\cmidrule(lr){3-3}
\cmidrule(lr){4-4}
\cmidrule(l){5-5}

\multirow{5}{*}{SciTS}
& Qwen3-8B
& 52.35
& 64.90 / 66.70 / 18.80 / \underline{66.20}
& zero-shot \\

& Gemini-2.5-Flash (Text)
& \underline{96.50}
& 60.90 / 64.80 / \underline{19.50} / 64.60
& zero-shot \\

& DeepSeek-V3
& 76.29
& 59.30 / 50.70 / 6.40 / 64.70
& zero-shot \\

& TimeOmni-SciTS
& \best{100.00}
& \underline{65.20} / \best{92.70} / \best{23.00} / 64.80
& SciTS-trained \\

& \textbf{\method{}}
& \best{100.00}
& \best{67.30} / \underline{90.19} / 18.78 / \best{67.39}
& zero-shot \\

\bottomrule
\end{tabular}
\caption{\textbf{Protocol-level external evaluation.}
TSQA reports forecasting and imputation MSE (F/I) and classification and
anomaly accuracy (C/A), using 1,000 adaptation examples per task. IRTS
reports overall, anomaly-detection (A.D.), classification (CLS), and
temporal-relation (TR) scores. ARF reports its official overall score,
Tier~I accuracy, and weighted F1. SciTS numerical SR is computed over all
4,019 numerical examples, and anomaly results report F1 on MEU01, PHU04,
PHU05, and URU04. Bold and underline denote the best and second-best
distinct values within each benchmark block. Results are not pooled across
protocols.}
\label{tab:external_summary}
\end{table*}

Under supervised adaptation, \method{} achieves the strongest displayed
TSQA-50 results on forecasting, imputation, classification, and anomaly
detection. Using the unchanged central checkpoint, it obtains the highest
displayed IRTS-ToolBench overall and temporal-relation scores and improves
all reported ARFBench metrics over the displayed baselines. These results
indicate that the compiled task representations remain useful under task
definitions and evaluation procedures not matched to the \exectsqa{}
contracts.

On SciTS, \method{} achieves 100\% numerical success without SciTS-specific
adaptation. Numerical success measures whether a valid numerical response is
produced and should not be interpreted as numerical prediction accuracy. For
anomaly detection, \method{} obtains the highest F1 on MEU01 and URU04,
ranks second on PHU04, and remains weaker on PHU05. Complete SciTS
classification results, numerical errors, balanced accuracies, and task-level
metrics are reported in the supplement.

Overall, the external results support transfer across supervised adaptation,
zero-shot irregular-series reasoning, anomaly-oriented question answering,
and scientific time-series analysis. They do not imply uniform predictive
superiority, since performance remains dependent on the task, dataset, and
evaluation protocol.

\subsection{Mechanism and Ablation Studies}

\begin{figure}[!t]
\centering
\IfFileExists{DecisionExecution-Mechanism.pdf}{
\includegraphics[width=0.82\columnwidth]{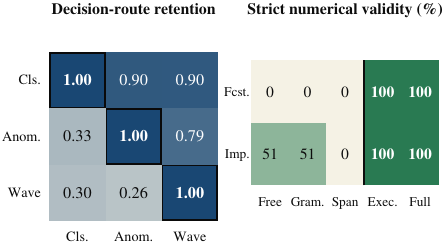}
}{
\fbox{\parbox[c][0.16\textheight][c]{0.98\columnwidth}{\centering
\textbf{Figure placeholder: DecisionExecution-Mechanism.pdf.}}}
}
\caption{\textbf{Reliability by construction and decision-path
specialization.}
The numerical panel reports contract-valid coverage as output control
progresses from free decoding to grammar constraints, structured spans, and
task-native construction. The decision panel compares shared and specialized
language paths.}
\label{fig:decision_execution}
\end{figure}

We next examine whether reliability arises from task-native construction
rather than surface constraints, and whether executor and language-path
specialization contribute distinct forms of predictive utility.
Figure~\ref{fig:decision_execution} studies the output-construction ladder
and decision-path specialization, while Table~\ref{tab:sharing_study}
controls how executors and LoRA paths are shared. Additional temporal-evidence
ablations are reported in the supplement.

\begin{table}[!t]
\centering
\scriptsize
\renewcommand{\arraystretch}{1.07}
\setlength{\tabcolsep}{1.8pt}
\begin{tabular}{@{}lcccc@{}}
\toprule
\textbf{Configuration}
& \makecell{\textbf{Fcst.}\\MAE $\downarrow$}
& \makecell{\textbf{Imp.}\\MAE $\downarrow$}
& \makecell{\textbf{Cls.}\\Acc. $\uparrow$}
& \makecell{\textbf{Anom.}\\F1 $\uparrow$} \\
\midrule

Shared LoRA + shared exec.
& .381 & .461 & .623 & \underline{.814} \\

Shared LoRA + task exec.
& \best{.322} & \best{.423} & \underline{.635} & .000 \\

Task LoRA + task exec.
& \underline{.337} & \underline{.427}
& \best{.706} & \best{.847} \\

\bottomrule
\end{tabular}
\caption{\textbf{Controlled specialization under compiled task paths.}
Task-specific executors improve continuous prediction under a shared
language path, while task-specific LoRA paths restore decision-task
performance.}
\label{tab:sharing_study}
\end{table}

Figure~\ref{fig:decision_execution} distinguishes syntactic control from
task-object reliability. Free and grammar-constrained decoding produce no
contract-valid forecasting tensors and reach only 51\% imputation coverage.
Structured spans identify where a numerical answer should appear, but do not
determine tensor rank, prediction horizon, channel organization, scale, or
missing-position alignment.

Task-native executors raise strict numerical validity to 100\%. This
progression shows that reliability does not follow automatically from
parseable syntax or a predefined answer span. It emerges when the compiled
horizon, mask, channel structure, scale state, and task contract directly
control construction of the numerical object.

Table~\ref{tab:sharing_study} separates executor specialization from
decision-path specialization. Replacing the shared executor with
task-specific executors reduces forecasting and imputation MAE from
.381/.461 to .322/.423 under the same shared LoRA path, but collapses
anomaly F1 to zero. Introducing task-specific LoRA paths restores
classification accuracy to .706 and anomaly F1 to .847.

The two mechanisms therefore serve complementary roles. Task-native
numerical executors preserve the inductive structure required for continuous
objects, whereas specialized language paths preserve task-dependent decision
behavior. Neither component alone provides balanced utility across continuous
and decision tasks, supporting the complete compiler-executor design.

\paragraph{Compilation from visible requests.}

The selector correctly compiles every forecasting, imputation,
classification, and anomaly-detection request into its corresponding path,
together with 530 of 559 waveform-analysis requests, yielding 99.40\% route
accuracy. The 29 unsuccessful cases consist of one runtime failure and 28
waveform-routing failures. No explicit task-type identifier or
evaluator-side route correction is used. The concentration of failures in
waveform analysis reflects the greater heterogeneity of its structured
requests, rather than a general inability to distinguish the five task
families.

Additional component studies show that the Wave Adapter reduces forecasting
and imputation MAE, whereas the route-conditioned resampler contributes most
strongly to anomaly detection. Their effects remain task dependent: removing
the Wave Adapter improves classification, while a linear projector obtains
the highest waveform-reasoning score. These results suggest that raw-wave
evidence and task-aware compression contribute complementary inductive
biases rather than uniformly improving every task. The complete configuration
is therefore selected for balanced predictive utility across all five task
families rather than optimal performance on every individual metric.
\section{Conclusion}

We identify \emph{task-object hallucination} as a reliability problem in
time-series language modeling: a plausible response may fail to instantiate
the numerical tensor, legal decision, or structured record required by its
task. \exectsqa{} measures this failure by separating contract-valid coverage
from predictive quality. We propose \method{}, a compiler-executor model
that compiles requests and wave-grounded evidence into typed task states and
constructs task-native outputs through task-specific paths. A single
checkpoint achieves 99.40\% contract-valid coverage across five task
families, versus 37.83\% for the strongest string-first baseline, while
retaining utility across four benchmarks and strong cross-benchmark transfer
evidence. These results establish task compilation as a route to reliable
time-series outputs.

\bibliography{aaai2027}
\end{document}